\documentclass[conference,letterpaper]{IEEEtran} 

\IEEEoverridecommandlockouts

\usepackage[
    letterpaper, 
    left=54pt,   
    right=54pt,  
    top=72pt,    
    bottom=54pt, 
]{geometry}

\usepackage{cite}
\usepackage{amsmath,amssymb,amsfonts}
\usepackage{algorithmic}
\usepackage{tabularx}
\usepackage{graphicx}
\usepackage{textcomp}
\usepackage{xcolor}
\usepackage{url} 
\usepackage{multirow}
\usepackage{enumitem}
\usepackage{booktabs}    
\usepackage{caption}     

\def\BibTeX{{\rm B\kern-.05em{\sc i\kern-.025em b}\kern-.08em
    T\kern-.1667em\lower.7ex\hbox{E}\kern-.125emX}}

\begin{document}

\title{Bridging Human Oversight and Black-box Driver Assistance: Vision-Language Models for Predictive Alerting in Lane Keeping Assist Systems\\}

\author{%
    \IEEEauthorblockN{%
        \begin{tabular}{@{}c@{\hskip 1.5cm}c@{}}%
            Yuhang Wang &
            Hao Zhou\textsuperscript{*}\thanks{*\,Hao Zhou is the corresponding author (email: \texttt{haozhou1@usf.edu}).} \\[2pt]
            \small yuhangw@usf.edu &
            \small haozhou1@usf.edu
        \end{tabular}%
    }\\[4pt]
    \IEEEauthorblockA{%
        \textit{Department of Civil \& Environmental Engineering},\\%
        \textit{University of South Florida}, Tampa, FL, USA%
    }%
}

\maketitle


\begin{abstract}

Lane Keeping Assist (LKA) systems, while increasingly prevalent, often suffer from unpredictable real-world failures, largely due to their opaque, black-box nature, which limits driver anticipation and trust. To bridge the gap between automated assistance and effective human oversight, we present \textbf{LKAlert}, a novel supervisory alert system that leverages Vision-Language Model (VLM) to forecast potential LKA risk 1–3 seconds in advance. LKAlert processes dash-cam video and CAN data, integrating surrogate lane segmentation features from a parallel interpretable model as automated guiding attention. Unlike traditional binary classifiers, LKAlert issues both predictive alert and concise natural language explanation, enhancing driver situational awareness and trust. To support the development and evaluation of such systems, we introduce \textbf{OpenLKA-Alert}, the first benchmark dataset designed for predictive and explainable LKA failure warnings. It contains synchronized multimodal inputs and human-authored justifications across annotated temporal windows. We further contribute a generalizable methodological framework for VLM-based black-box behavior prediction, combining surrogate feature guidance with LoRA. This framework enables VLM to reason over structured visual context without altering its vision backbone, making it broadly applicable to other complex, opaque systems requiring interpretable oversight. Empirical results correctly predicts upcoming LKA failures with 69.8\% accuracy and a 58.6\% F1-score. The system also generates high-quality textual explanations for drivers (71.7 ROUGE-L) and operates efficiently at approximately 2 Hz, confirming its suitability for real-time, in-vehicle use. Our findings establish LKAlert as a practical solution for enhancing the safety and usability of current ADAS and offer a scalable paradigm for applying VLMs to human-centered supervision of black-box automation.

\end{abstract}




\section{Introduction}
\label{sec:introduction}

Advanced Driver Assistance Systems (ADAS), particularly Level-2 features like Lane Keeping Assist (LKA), are increasingly prevalent in modern vehicles, offering significant potential to enhance driving safety and comfort. However, the real-world operational capabilities of current LKA systems often fall short of expectations. As highlighted by recent analyses and datasets \cite{wang2025openlka}, these systems frequently encounter difficulties when faced with common environmental and infrastructural variations, such as poorly maintained lane markings, complex road geometries, temporary occlusions, or adverse weather conditions. These challenges can lead to unexpected performance degradation or outright system failure, potentially compromising safety.

The critical issue arising from these limitations is the unpredictability of LKA failures from the driver's perspective. The complex interplay between the system's perception, decision-making, and control modules makes it exceedingly difficult for a typical driver to anticipate when the LKA might struggle or disengage. While subtle cues might occasionally foreshadow trouble, relying on drivers to consistently detect and interpret these precursors is neither reliable nor safe. This unpredictability creates a dangerous situation where drivers may over-trust the system, leading to potential accidents, or conversely, under-utilize the technology due to lack of confidence, thereby negating its intended safety benefits. Consequently, a distinct gap exists: the lack of a reliable mechanism to proactively inform the driver before LKA performance is compromised.

To bridge this critical gap between automated assistance and effective human oversight, we propose \textbf{LKAlert}, a novel supervisory alert system leveraging Vision-Language Models (VLMs) to forecast potential LKA failures 1–3 seconds in advance based on dash-cam video and CAN data. Crucially, LKAlert enhances driver situational awareness, trust, and response effectiveness by providing not only a timely predictive alert but also a concise natural language explanation—addressing the opacity of current systems. Development is supported by our introduction of the \textbf{OpenLKA-Alert} dataset, the first benchmark specifically curated for such predictive and explainable warning systems. Methodologically, LKAlert fuses multimodal inputs with interpretable spatial guidance derived from lane segmentation masks, utilizing efficient parameter-efficient VLM adaptation to achieve context-aware reasoning and improve the safety and usability of existing LKA technology.

Our core contributions are:

\begin{itemize}[leftmargin=1.2em]
\item \textbf{OpenLKA-Alert Dataset:} We curate and release OpenLKA-Alert, the first benchmark dataset purpose-built for developing and evaluating LKA failure prediction systems that are both predictive and explainable. Its provision of synchronized multimodal data alongside natural language justifications fills a critical resource gap, enabling future research into more transparent and trustworthy ADAS safety features.

\item \textbf{LKAlert System:} We propose \textbf{LKAlert}, a novel early warning framework that leverages Vision-Language Models (VLMs) to predict LKA disengagements 1–3 seconds in advance. To our knowledge, LKAlert is the first supervisory alert system that combines dash-cam video and large language models to enhance driver awareness and oversight of black-box ADAS features. By fusing multimodal inputs—visual scenes, surrogate lane structure, and vehicle dynamics—LKAlert produces both timely alerts and natural language explanations, facilitating safer and more informed driver intervention.

\item \textbf{A Generalizable Framework for VLM-based Black-box Prediction:} We introduce and validate a methodological framework that enables VLMs to predict the behavior of complex, opaque systems like commercial LKA modules. It integrates two core strategies:
(1) \textbf{Surrogate Feature Guidance:} An open-source interpretable model (LaneNet) provides explicit lane segmentation masks, which serve as spatially grounded, task-relevant surrogate features for the VLM.
(2) \textbf{Decoder-Only Adaptation:} We fine-tune only the language decoder while keeping the vision encoder frozen, enabling efficient training that focuses the model on learning how to interpret the surrogate spatial guidance.
Together, these strategies yield high predictive accuracy and explanation quality (+ 12.4\% F$_1$, +68.2 ROUGE-L), offering a scalable and lightweight paradigm for deploying VLMs in real-time safety-critical applications involving black-box automation in ADAS and beyond.

\end{itemize}

\section{Related Work}
\label{sec:related_work}

Advanced driver assistance systems (ADAS) have played a crucial role in mitigating road accidents and enhancing overall traffic safety. Studies have demonstrated that features such as lane keeping assist (LKA) reduce crash frequency~\cite{spicer2021effectiveness,mansourifar2025novice}. However, a number of issues persist in real-world deployments. Numerous works have identified that inadequate lane marking visibility—especially on rural roads under low daylight or adverse weather conditions—can lead to misinterpretations in lane merging and diverging scenarios, ultimately causing erroneous vehicle positioning or system disengagement~\cite{ding2024exploratory,masello2022road}. These issues are compounded in situations where multiple adverse factors, such as complex traffic markings and high-curvature bends, coexist. As a result, current evaluation metrics often fall short of capturing the multifaceted challenges that affect LKA performance in the field. Such limitations underscore the pressing need for innovative approaches that not only predict LKA failures under diverse conditions but also offer interpretable feedback on the underlying causes.

Vision-language models (VLMs) have transformed the way we understand and process multimodal information by learning to align visual and textual representations. Models such as Qwen2.5-VL-7B-Instruct, BLIP, and CLIP leverage large-scale pre-training on image-text pairs to yield robust, semantically aligned features. Significant enhancements in regional understanding have been achieved through frameworks like RegionGPT, which makes simple yet effective modifications to boost spatial awareness in the vision encoder~\cite{guo2024regiongpt}. Parallel efforts in developing lightweight models, such as MiniVLM~\cite{wang2020minivlm}, have demonstrated that fast inference can be achieved without incurring a significant performance penalty. Moreover, recent research has also explored methods to adapt large VLMs for resource-constrained edge environments~\cite{cai2024self}, emphasizing the need to balance model complexity with real-world deployment feasibility.

LoRA injects trainable low-rank matrices into frozen Transformer projections, allowing fewer than 1 \% of the weights to be updated while matching full fine-tuning accuracy~\cite{hu2022lora}.  Recent refinements such as CLIP-LoRA for few-shot adaptation~\cite{zanella2024low} and LoRA$^{+}$ for rank-balanced optimization~\cite{hayou2024lora+} further improve convergence and stability, a trend surveyed across visual PEFT techniques~\cite{xu2023parameter}.  In the autonomous-driving domain, DriveLLaVA employs LoRA to align large vision–language models with human-level behaviour decisions~\cite{zhao2024drivellava}, while CarLLaVA leverages the same strategy to attain state-of-the-art camera-only closed-loop control in CARLA~\cite{renz2024carllava}.  LoRA-adapted CLIP backbones deliver a 91 \% F\textsubscript{1} score on driving-scene classification and operate in real time for ADAS pipelines~\cite{elhenawy2025vision}.  Lightweight hazard-detection frameworks likewise show that LoRA lowers compute costs without sacrificing multimodal reasoning~\cite{shen2024multimodal}.  Edge-deployment studies confirm that pairing LoRA with quantization meets the stringent latency and power budgets of in-vehicle GPUs~\cite{zhang2025language}.  Benchmarks such as CODA-LM further validate that LoRA-enhanced VLMs maintain robustness on self-driving corner cases~\cite{chen2024driving}. 

In the field of autonomous driving, the integration of VLMs emerges as a promising solution to address the limitations of conventional ADAS. VLMs have been used to analyze complex driving scenes and generate high-quality pre-annotations that facilitate human evaluation~\cite{chen2025automated}. Surveys indicate that these models hold substantial potential to overcome challenges associated with ambiguous lane markings and intricate traffic patterns~\cite{zhou2024vision,cui2024survey}. End-to-end frameworks such as VLP~\cite{pan2024vlp} and DriveGPT4~\cite{xu2024drivegpt4} illustrate the benefits of planning and reasoning in challenging scenarios. Structured systems like Senna~\cite{jiang2024senna} and evaluations of safety cognition in VLMs~\cite{zhang2025evaluation} further reveal that combining detailed visual semantics with natural language explanations not only improves interpretability but also enhances overall system robustness. These developments point toward the necessity for novel datasets and innovative model designs that fully exploit the multimodal nature of driving data, setting the stage for our work using Qwen2.5-VL-7B-Instruct as the backbone for an interpretable LKA Alert System.


\section{Methodology}
\label{sec:methodology}

\subsection{Data Preparation}
\label{sec:dataset}

\begin{figure}[!h]
  \centering
  \includegraphics[width=0.48\textwidth]{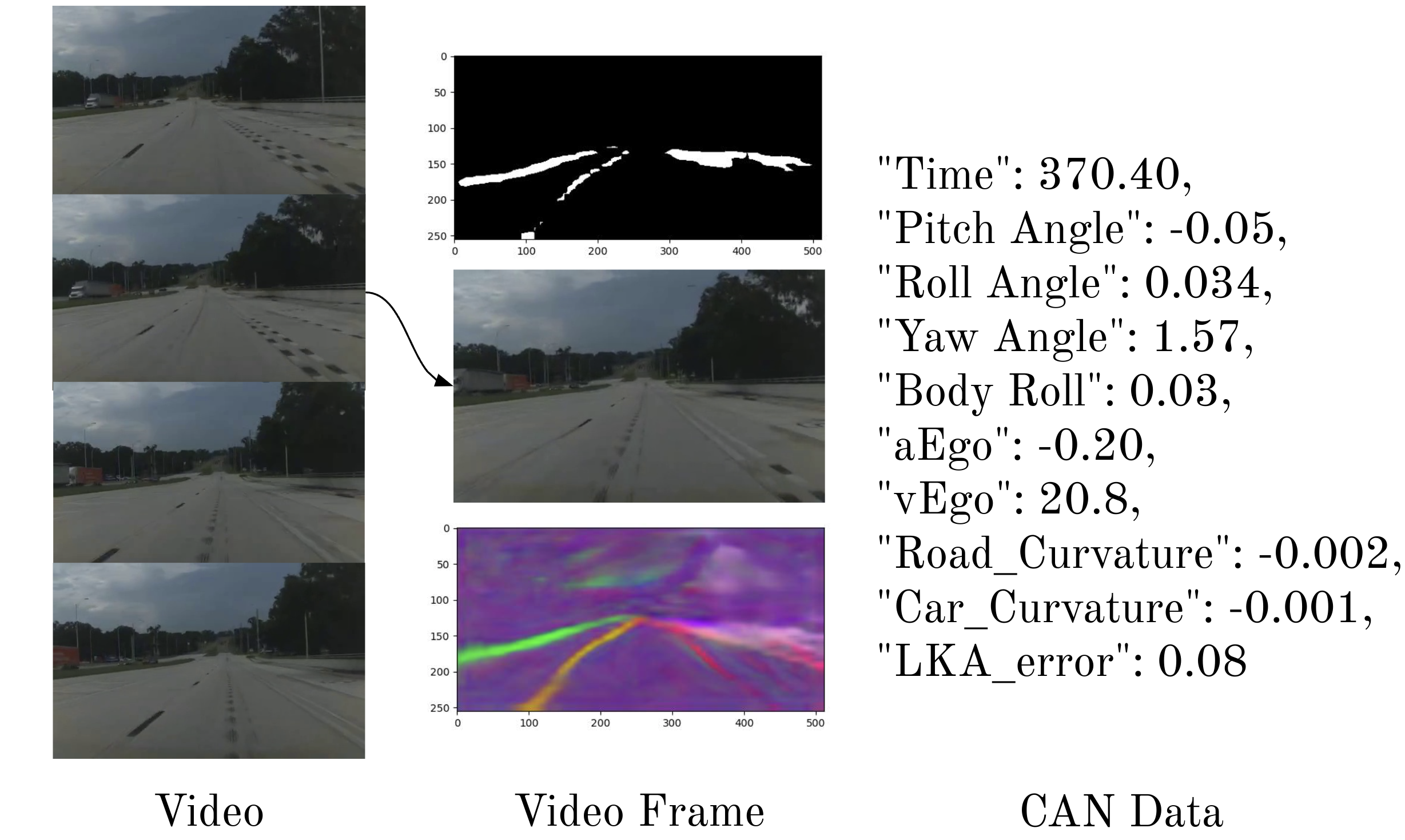}
  \caption{The video file is segmented into video frames every 0.5 second and preprocessed by Lanenet to obtain the intermediate process of lane line recognition: Binary Segment and Instance Segment, which provide interpretability for edge segmentation and segmentation of different lane lines respectively. The vehicle control information in CAN and the output information of Openpilot are used as text data, which can be provided together with the previous visual data during training.}
  \label{fig:dataset_oaga}
\end{figure}

We generate the OpenLKA-Alert dataset required for supervised training based on the OpenLKA-Failure dataset and the OpenLKA-Normal dataset. OpenLKA-Failure and OpenLKA-Normal are obtained based on the OpenLKA dataset\cite{wang2025openlka} . OpenLKA-Normal was obtained by sampling along with the normal driving scenarios and was used to continue balancing with the data from OpenLKA-Failure. Based on the OpenLKA driving video, CAN and Openpilot outputs, OpenLKA-Failure combines the vehicle position information from CAN and Openpilot, LKA state information, vehicle attitude information and the model outputs of the vehicle's position in the lane to locate the LKA facing a larger Lane Centering Deviation and LKA disengagement. The specific location of the LKA disengagement is obtained by extending it according to a window of 3.5 seconds forward and 2.5 seconds backward. Thus OpenLKA-Failure not only contains detailed video and CAN of the moment when the LKA experienced a large Lane Centering Deviation or disengagement, but also contains information for a total of 6-second windows before and after the problem occurred. This will help us to analyse the cause of the LKA problem and design the OpenLKA-Alert dataset based on it. This dataset is designed as follows:

As shown in Fig.~\ref{fig:dataset_oaga}, we first sample all the video data within 0.5 seconds and convert the video data into video frame images. Then we retrieve the corresponding CAN information according to the time to extract the detailed vehicle control information.

\begin{figure}[!h]
  \centering
  \includegraphics[width=0.48\textwidth]{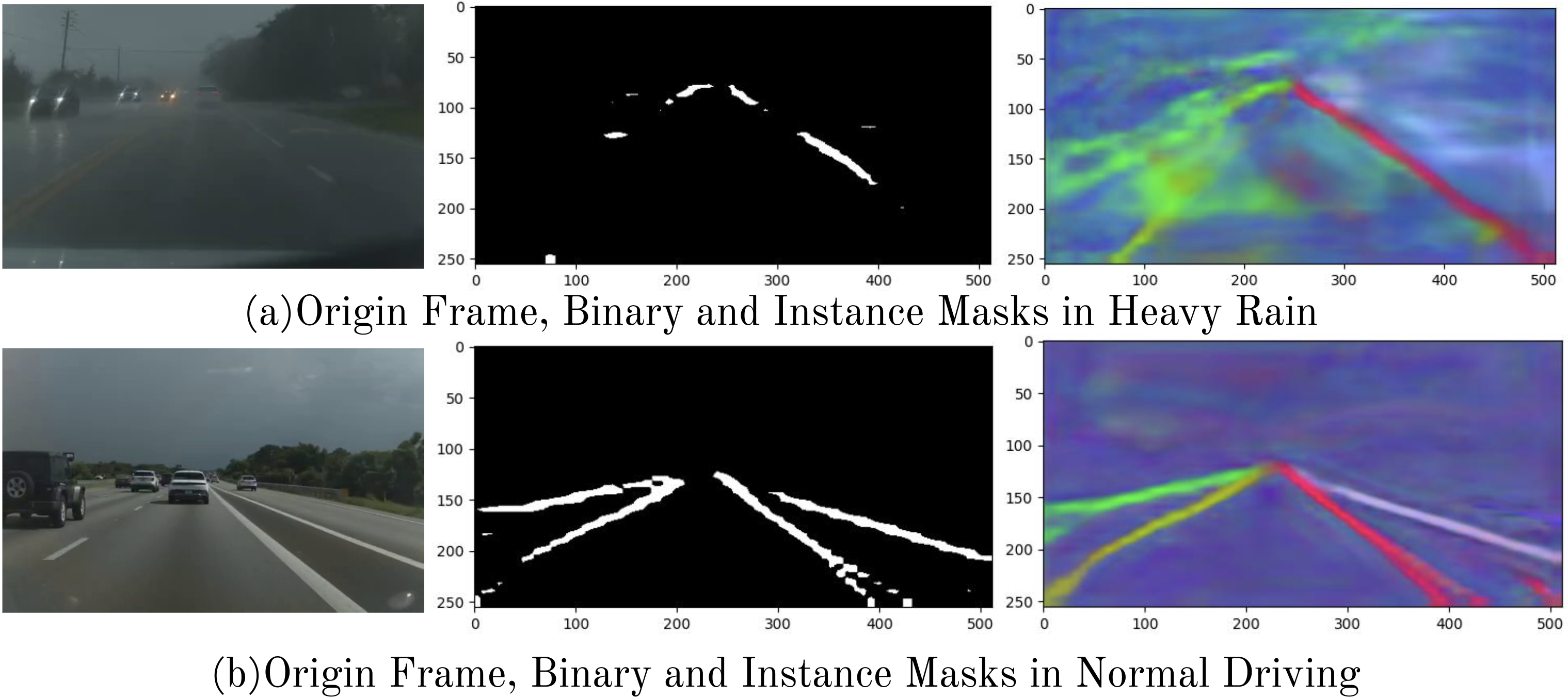}
  \caption{Visual Data From OpenLKA-Failure Dataset and OpenLKA-Normal Dataset}
  \label{fig:VisualDataExample}
\end{figure}

Then, we use LaneNet\cite{neven2018towards} to preprocess each image frame, which is used to simulate the road marking detection algorithm of the vehicle system, and we extract the Binary segments and Instance Segments in the detection results as the results of lane line edge segmentation and lane line instance segmentation, which provide intermediate interpretations for the final lane line detection. The reason we use LaneNet is not only that it is an effective road line detection method, but more importantly, it can provide interpretable masks. In addition, the reason why we do not use the SOTA road line detection method is that many of the car models in OpenLKA are from 2019 to 2022. For the car systems of this period, if the SOTA method is used, their capabilities will be overestimated, which is not good for designing the Alert System from the perspective of traffic safety. Fig.~\ref{fig:VisualDataExample} shows two examples in our OpenLKA-Alert dataset. The left part, the binary mask indeicates a failure to divide the lanelines in the near future, and in the instance mask, the left lanelines and the road edge are divided as well, showing the ADAS system can nearly detect lanelines correctly in the heavy rain, causing a safety risk. However, in the right part, based on the binary and instance segments, all of the lanelines and road edges are successfully segmented and detected, indicates that the vehicle is running normally.

Since the scene where the LKA problem occurs is fixed to the 7th frame of the video frame image, we have performed a lot of manual screening on each extracted OpenLKA-Failure image. We will only retain the current image if it has the same features as the 7th frame image that will cause LKA failure. Otherwise, we remove this image from the data. We finally got an OpenLKA-Alert dataset that contains a continuous sequence of video data and CAN data from the time when the vehicle LKA is about to have a problem to the time when the LKA has already had a problem.

\begin{figure}[!h]
  \centering
  \includegraphics[width=0.48\textwidth]{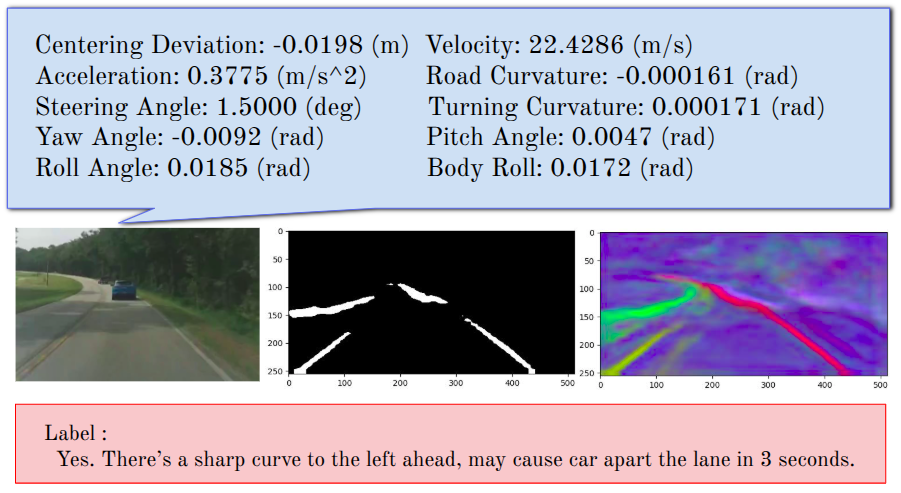}
  \caption{Although the current laneline is clear, there is a sharp curve ahead, and the car will lean to the right, almost depart the lane in 3 seconds, so this data is marked as \textbf{Alert.}}
  \label{fig:alert_example}
\end{figure}

We annotate all the image frames of OpenLKA-Failure and OpenLKA-Normal with whether or not we should let the Alert System do the alerting for the current image, and we always annotate the Failure dataset with ‘Yes’ and the Normal dataset with ‘No.’ In addition, for the OpenLKA-Failure dataset, we also annotate the reason why we need the alert system to further provide rich interpret-ability. As shown in Fig.~\ref{fig:alert_example}, although the lane detection of the vehicle's current ADAS system is normal and the vehicle is driving normally, it can be seen that it is about to enter a curve with a large curvature. Therefore, we still believe that an early warning is needed in this frame, and this data is marked as "Yes. There’s a sharp curve to the left ahead, which may cause the car to apart the lane in 3 seconds.".

\begin{figure}
  \centering
  \includegraphics[width=0.45\textwidth]{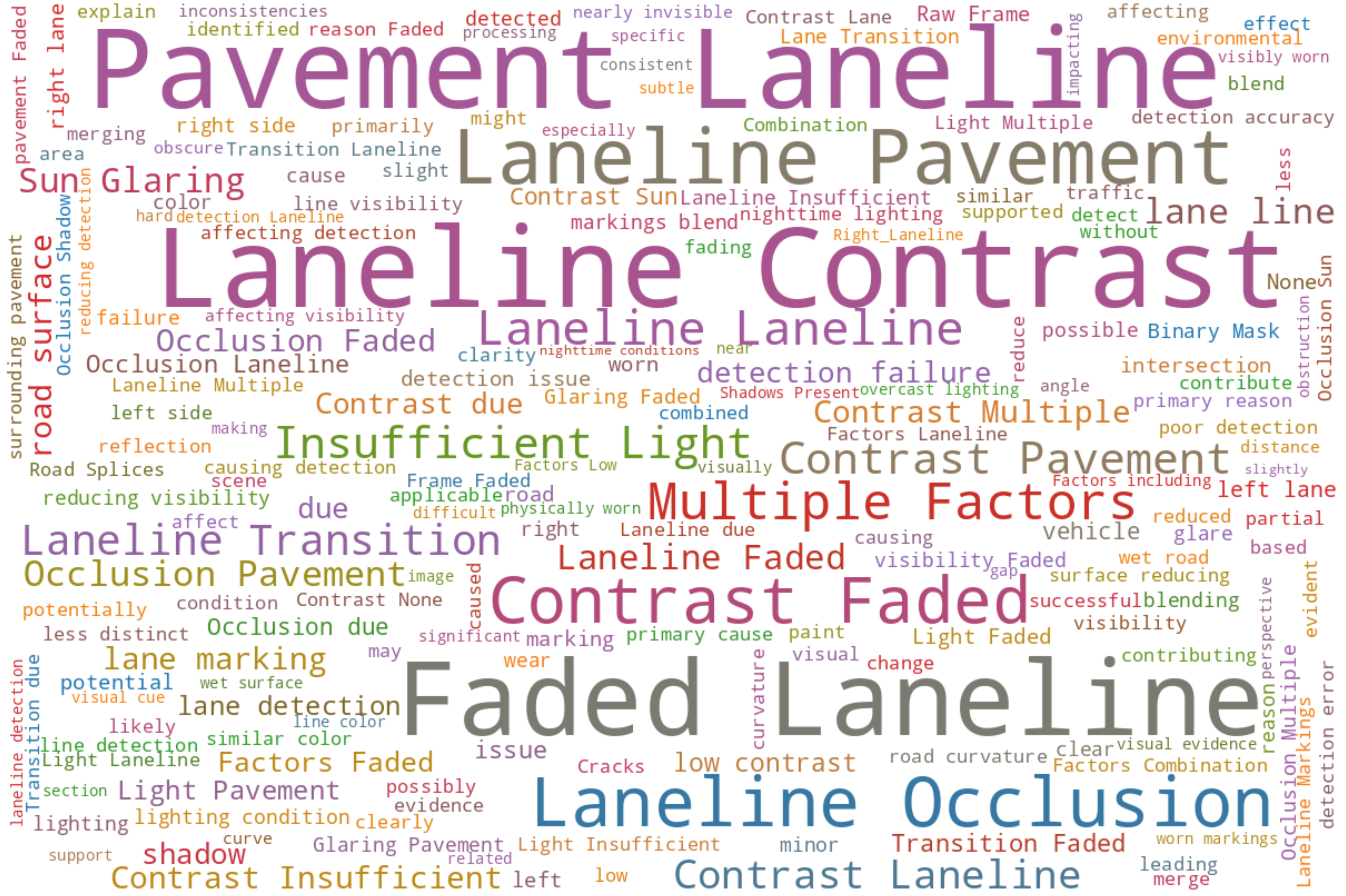}
  \caption{Semantic word cloud of the LKA Failure Explanation in OpenLKA-Alert}
  \label{fig:world_cloud}
\end{figure}

We summarize the features of the LKA problems we annotated as Fig.~\ref{fig:world_cloud}. From this word cloud, we can also conclude that the common challenges of LKA are mainly due to faded laneline, laneline occlusion, and low contrast between pavement and laneline.

\subsection{Model Structure}
\label{sec:alert_system}

\begin{figure*}[!ht]
  \centering
  \includegraphics[width=0.95\textwidth]{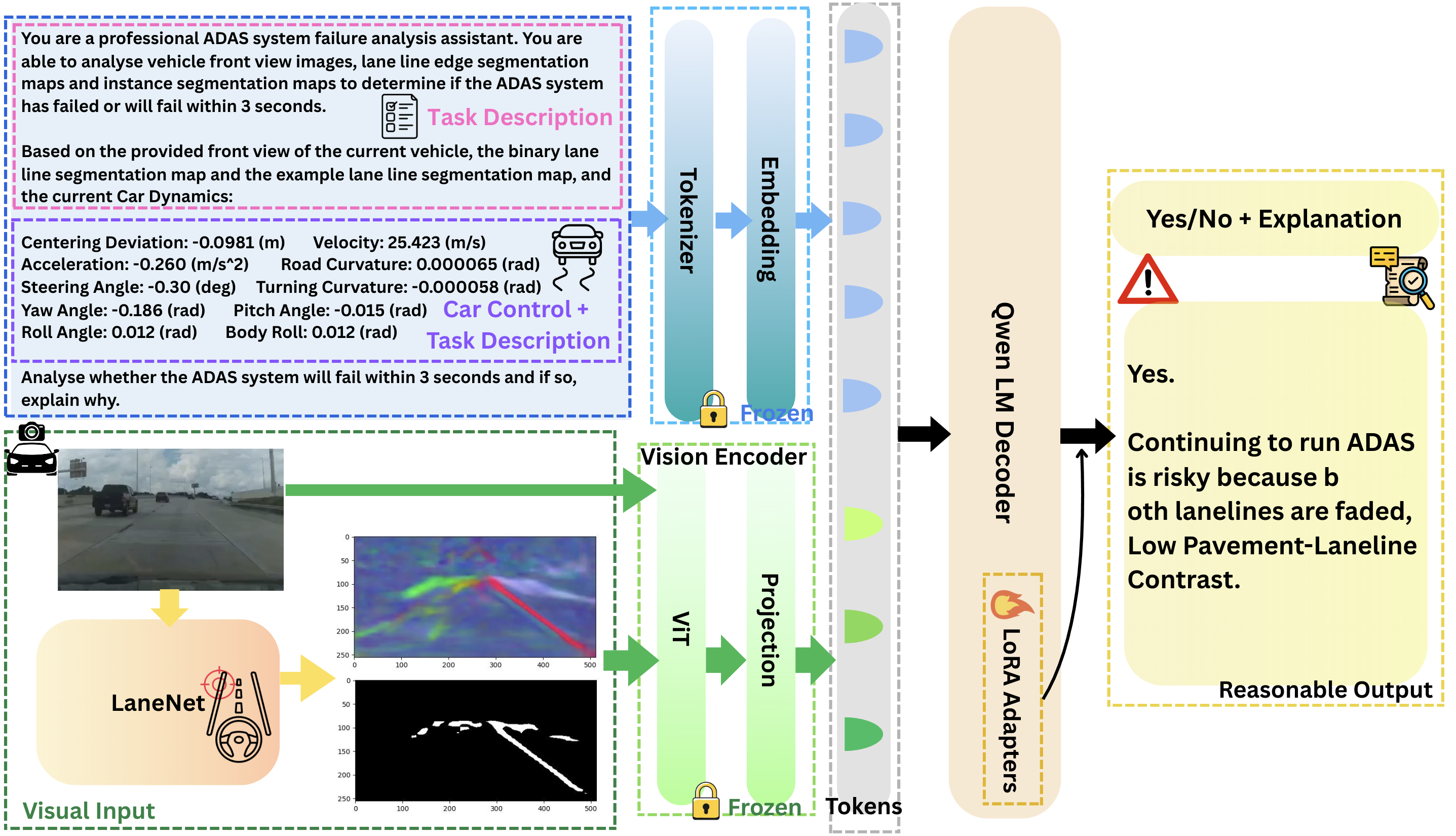}
  \caption{LKAlert Architecture: Multimodal inputs (Image $\mathcal{I}_{rgb}$, Masks $\mathcal{M}_{bin}, \mathcal{M}_{ins}$, CAN $\mathbf{c}$) are processed by frozen encoders ($f_{ViT}, f_{Text}$). The combined representation $\mathbf{X}$ feeds the LoRA-adapted decoder $f_{Dec}$ to generate the alert $y$ and explanation $e$.}
  \label{fig:Model}
\end{figure*}

LKAlert predicts imminent Lane Keeping Assist (LKA) failures and provides explanations, formulated as a conditional generation task.

\subsubsection{Inputs and Outputs}
The model input is a multimodal set $\mathcal{S} = \{\mathcal{I}_{rgb}, \mathcal{M}_{bin}, \mathcal{M}_{ins}, \mathbf{c}\}$, containing the RGB image, CAN dynamics, and crucial LaneNet~\cite{neven2018towards}-generated binary ($\mathcal{M}_{bin}$) and instance ($\mathcal{M}_{ins}$) lane segmentation masks. These masks provide explicit geometric guidance. The target output is a sequence $O = (y, e)$, where $y \in \{\text{Yes}, \text{No}\}$ is the failure alert and $e$ is the textual explanation.

\subsubsection{Architecture Overview}
As shown in Fig.~\ref{fig:Model}, LKAlert utilizes a VLM (Qwen2.5-VL~\cite{bai2025qwen2}) structure. Frozen encoders ($f_{ViT}, f_{Text}$) process the input set $\mathcal{S}$ (with $f_{ViT}$ handling the image $\mathcal{I}_{rgb}$ alongside masks $\mathcal{M}_{bin}, \mathcal{M}_{ins}$) into a fused latent representation $\mathbf{X}$:
\begin{equation}
    \mathbf{X} = \text{Encode}(\mathcal{S}; \Theta_{frozen})
    \label{eq:simplified_encode}
\end{equation}
This representation $\mathbf{X}$ contains embedded visual appearance, vehicle dynamics, and explicit lane geometry features. A Language Model Decoder ($f_{Dec}$), adapted using LoRA, then generates the output $O$ autoregressively based on $\mathbf{X}$:

\begin{equation}
    O = f_{Dec}(\mathbf{X}; \Theta_{Dec,0} + \Delta\Theta_{LoRA})
    \label{eq:simplified_decode}
\end{equation}

\subsection{LoRA Fine-Tuning with Mask Guidance}
\label{sec:lora}

We fine-tune only the decoder of LKAlert using LoRA~\cite{hu2022lora}, adapting its pre-trained parameters $\Theta_{Dec,0}$ with low-rank updates $\Delta\Theta_{LoRA}$.

\subsubsection{Learning Objective}
The trainable LoRA parameters $\Delta\Theta_{LoRA}$ are optimized to maximize the conditional log-likelihood of the ground-truth alert $y^*$ and explanation $e^*$ from the OpenLKA-Alert dataset $\mathcal{D}$, given the fused input representation $\mathbf{X}$:
\begin{equation}
    \max_{\Delta\Theta_{LoRA}} \mathbb{E}_{(\mathcal{S}, y^*, e^*) \sim \mathcal{D}} \left[ \log p( (y^*, e^*) \,|\, \mathbf{X}; \Theta_{Dec,0} + \Delta\Theta_{LoRA}) \right]
    \label{eq:concise_objective}
\end{equation}
where $\mathbf{X} = \text{Encode}(\mathcal{S}; \Theta_{frozen})$ encodes the image, masks, and CAN data via frozen encoders.

\subsubsection{Mechanism of Mask-Guided Adaptation}
This training strategy effectively leverages the provided inputs:

\textbf{Rich Feature Embedding:} The frozen $f_{ViT}$ embeds not only visual appearance but also the explicit lane geometry from the input masks ($\mathcal{M}_{bin}, \mathcal{M}_{ins}$) into the representation $\mathbf{X}$. This makes critical geometric information directly available to the decoder.
    
\textbf{Targeted Decoder Learning:} By optimizing Eq.~\ref{eq:concise_objective}, the LoRA updates ($\Delta\Theta_{LoRA}$) specifically adapt the decoder to recognize correlations between the embedded features in $\mathbf{X}$ (including the geometric cues from masks) and the target LKA failure events ($y^*, e^*$).

Essentially, LoRA efficiently teaches the decoder to utilize the relevant geometric features—pre-processed by the frozen encoder from the input masks—that are indicative of the need for an LKA alert and its corresponding reason. This provides effective guidance towards the specific task without modifying or retraining the powerful, general-purpose frozen encoders. Key advantages include high parameter efficiency and zero additional inference latency, as LoRA matrices can be merged post-training.

\begin{table*}
\centering 
\caption{LKAlert performance before and after LoRA fine-tuning.}
\label{tab:loraprepost}
\begin{tabular}{lllrrrrrrrrr}
\toprule
Model & Accuracy & Precision & Recall & F1 & BLEU-4 & ROUGE-1 & ROUGE-2 & ROUGE-L & SPS \\
\midrule
Qwen2-VL-2B-origin & 52.80 & 45.96 & 19.96 & 27.83 & 3.86 & 10.32 & 1.84 & 5.39 & 0.65 \\
Qwen2-VL-2B-final & 62.70 & 71.81 & 29.61 & 42.17 & 14.39 & 31.04 & 1.40 & 25.57 & 2.85 \\
Qwen2.5-VL-3B-origin & 55.50 & 66.67 & 4.82 & 9.00 & 1.95 & 7.26 & 1.55 & 2.88 & 0.26 \\
Qwen2.5-VL-3B-final & 68.90 & 76.56 & 45.83 & 57.34 & 63.49 & 70.90 & 63.15 & 71.22 & 2.29 \\
Qwen2.5-VL-7B-origin & 47.60 & 43.44 & 49.34 & 46.20 & 2.57 & 7.37 & 1.40 & 3.57 & 0.24 \\
Qwen2.5-VL-7B-final & \textbf{69.80} & \textbf{78.02} & \textbf{46.71} & \textbf{58.63} & \textbf{64.09} & \textbf{71.39} & \textbf{63.99} & \textbf{71.72} & 1.97 \\
\bottomrule
\end{tabular}
\end{table*}

\subsection{Validation Methods}
\label{sec:val}

\textbf{Dataset.}  
A held-out set $\mathcal{D}_{\text{val}}$ of $1{,}000$ samples is drawn
uniformly from \textit{OpenLKA-Failure} and \textit{OpenLKA-Normal} that not contained in the training set
($456$ \textsf{Yes} and $544$ \textsf{No} cases).  
Each sample carries the same two-field annotation as the OpenLKA-Alert
training corpus: a warning label $y\!\in\!\{\text{Yes},\text{No}\}$ and
a reference explanation $e$.

\vspace{2pt}
\textbf{Decision reliability.}  
We tabulate
$\bigl(TP,FP,TN,FN\bigr)$ and report
\[
\text{Accuracy}=\frac{TP+TN}{|\mathcal{D}_{\text{val}}|},\qquad
\text{F-1}=\frac{2TP}{2TP+FP+FN},
\]
providing a direct safety proxy for the alert trigger.

\vspace{2pt}
\textbf{Explanation fidelity.}  
Natural-language quality is evaluated by
\textit{predict\_bleu-4} and \textit{predict\_rouge-$\{1,2,L\}$}:  
\begin{itemize}[leftmargin=1.2em,itemsep=1pt]
  \item \textbf{BLEU-4} gauges $n$-gram (\(n\!\le\!4\)) precision,
        rewarding fluent token overlap.
  \item \textbf{ROUGE-1/2/L} measure unigram, bigram and longest-common-sub\-sequence recall, 
        capturing semantic coverage of the ground-truth reason.
\end{itemize}

\vspace{2pt}
\textbf{Runtime efficiency.}  
For deployment realism we log \textit{predict\_samples\_per\_second}
\[
R_{\text{sps}}
=\frac{|\mathcal{D}_{\text{val}}|}{t_{\text{wall}}}\;
[\text{samples}\,\mathrm{s}^{-1}],
\]
where $t_{\text{wall}}$ is the wall-clock time to complete one validation
epoch on an RTX5090-32\,GB GPU.  
Together, these metrics quantify LKAlert’s safety effectiveness
(low $FN$), linguistic faithfulness (high BLEU/ROUGE) and real-time
viability (high $R_{\text{sps}}$).

\begin{table*}[!ht]
\centering
\footnotesize
\caption{Performance comparison of guided vs.\ unguided variants.}
\label{tab:lkalert_results}
\setlength{\tabcolsep}{3pt}
\begin{tabular}{lccccccccccccccc}
\toprule
Model & Setting &
Acc. & F1 & BLEU-4 & R1 & R2 & RL & SPS\,$\uparrow$ &
Acc\textsubscript{rank} & BLEU4\textsubscript{rank} & R1\textsubscript{rank} & R2\textsubscript{rank} & RL\textsubscript{rank} & SPS\textsubscript{rank} &
Avg.\,$\downarrow$ \\
\midrule
Qwen2-VL-2B      & Guided    & 62.70 & 42.17 & 14.39 & 31.04 & 01.40 & 25.57 & 2.85 & 5 & 6 & 5 & 6 & 6 & 4 & 5.33 \\
Qwen2.5-VL-3B    & Guided    & 68.90 & 57.34 & 63.49 & 70.90 & 63.15 & 71.22 & 2.29 & 2 & 2 & 2 & 2 & 2 & 5 & 2.50 \\
\textbf{Qwen2.5-VL-7B}    & Guided    & \textbf{69.80} & \textbf{58.63} & \textbf{64.09} & \textbf{71.39} & \textbf{63.99} & \textbf{71.72} & 1.97 & \textbf{1} & \textbf{1} & \textbf{1} & \textbf{1} & \textbf{1} & 6 & \textbf{1.83} \\
\midrule
Qwen2-VL-2B      & Unguided  & 61.10 & 36.75 & 15.32 & 30.99 & 03.32 & 25.73 & 5.96 & 6 & 5 & 6 & 5 & 5 & 1 & 4.67 \\
Qwen2.5-VL-3B    & Unguided  & 67.50 & 57.29 & 62.09 & 69.63 & 61.46 & 70.07 & 5.38 & 4 & 4 & 4 & 4 & 4 & 2 & 3.67 \\
Qwen2.5-VL-7B    & Unguided  & 67.70 & 55.93 & 62.42 & 69.82 & 61.98 & 70.29 & 4.60 & 3 & 3 & 3 & 3 & 3 & 3 & 3.00 \\
\bottomrule
\end{tabular}
\end{table*}

\section{Results}
\label{sec:results}

\subsection{Model Performance} 

A primary evaluation involves comparing the performance of the models before ('origin') and after ('final') LoRA fine-tuning, as detailed in Table~\ref{tab:loraprepost}. The results consistently show significant improvements across all model scales after adaptation. For instance, the Qwen2.5-VL-7B model exhibits substantial gains: prediction accuracy increases from 47.60\% to \textbf{69.80\%}, the F1-score improves from 46.20\% to \textbf{58.63\%}, and the explanation quality, measured by ROUGE-L, surges from a baseline of 3.57 to \textbf{71.72}. This trend confirms that the LoRA fine-tuning process effectively adapts the pre-trained VLMs to the specific demands of the LKAlert task, enhancing both failure prediction and justification generation capabilities.

Our best performing configuration is the fine-tuned Qwen2.5-VL-7B model incorporating mask guidance (represented as 'Qwen2.5-VL-7B-final' in Table~\ref{tab:loraprepost} and 'Guided' in Table~\ref{tab:lkalert_results}). This model achieves a final accuracy of 69.80\% and an F1-score of 58.63\% for the alert prediction task. The high ROUGE-L score of 71.72 indicates its proficiency in generating relevant natural language explanations for the predicted alerts.

\begin{figure}[!h]
  \centering
  \includegraphics[width=0.48\textwidth]{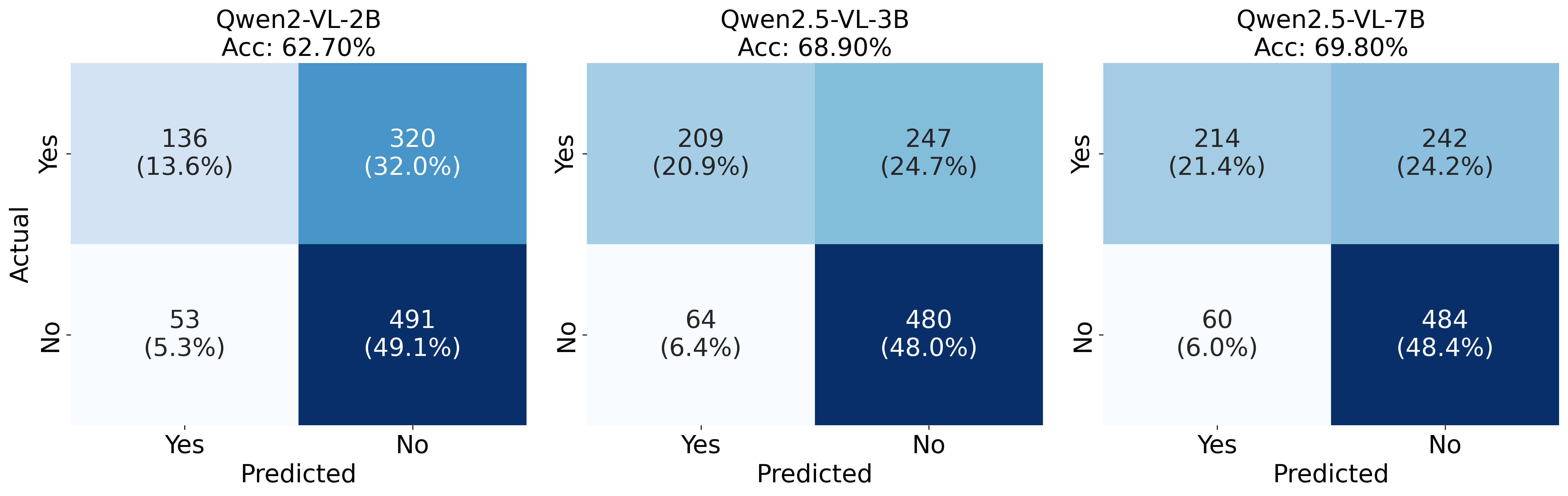}
  \caption{Confusion matrices of different models}
  \label{fig:ConfusionMatrix}
\end{figure}

Analyzing the confusion matrix for this final 7B model (Figure~\ref{fig:ConfusionMatrix}, right panel) provides further insight. As noted in the original analysis accompanying the figure, model scaling improves the identification of true failure events (True Positives constitute 21.4\% of the validation samples for the 7B model) and reduces missed detections (False Negatives are 24.2\%). While reducing false negatives remains paramount for safety-critical systems, the final 7B LKAlert model demonstrates a strong capability to anticipate potential LKA failures compared to its pre-trained state, effectively leveraging the learned adaptations and the guidance provided by the input masks.

\subsection{Fine-Tuning Convergence and Efficiency} 
\label{sec:training_process} 

\begin{figure}[!h]
 \centering
 \includegraphics[width=0.48\textwidth]{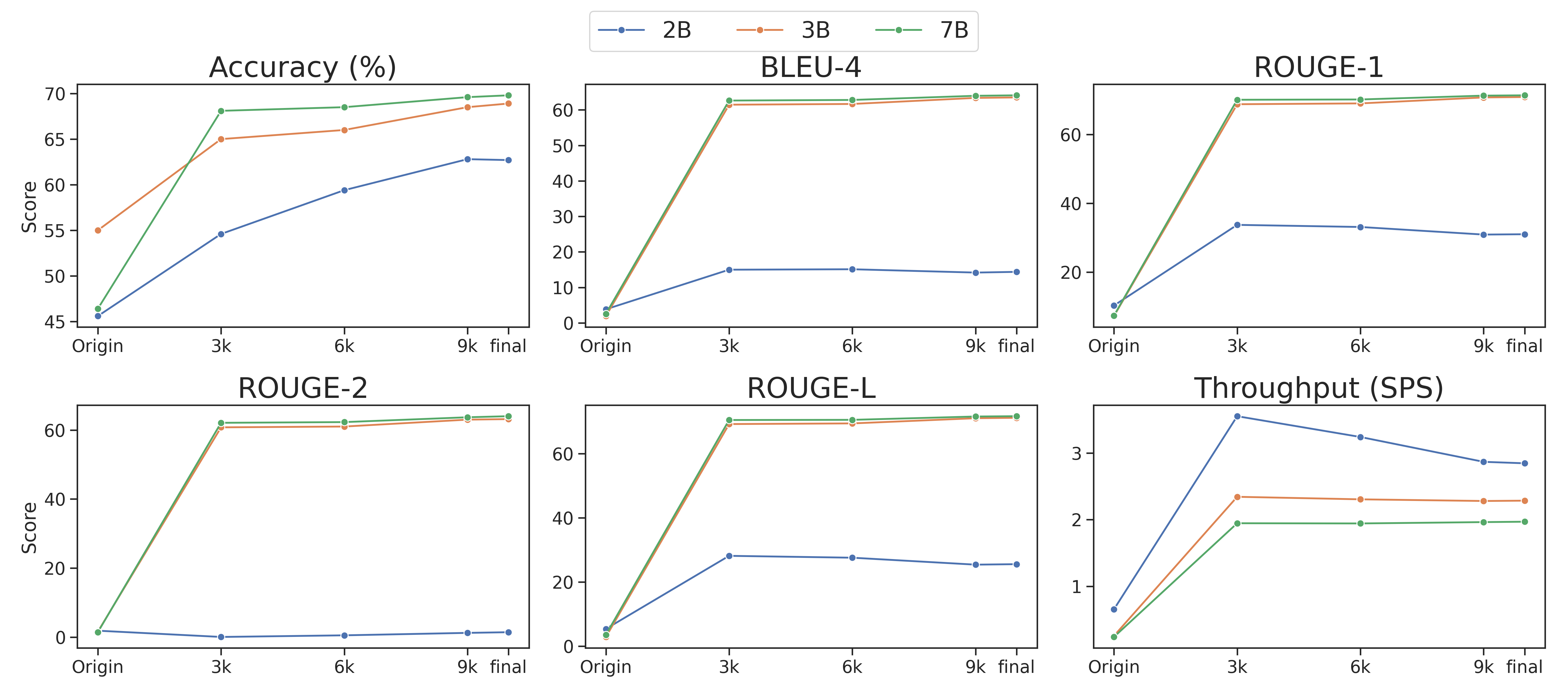} 
 \caption{The Validation Results of Different Models During Fine-tuning} 
 \label{fig:ResultPlot}
\end{figure}

Figure~\ref{fig:ResultPlot} plots the validation metrics against training steps, illustrating the LoRA fine-tuning dynamics for the different model sizes. We observe several key trends:

\begin{enumerate}[label=(\roman*),leftmargin=1.1em]
    \item \textbf{Rapid Language Metric Convergence:} For the explanation generation task, metrics like BLEU-4 and ROUGE scores show substantial improvement within the initial 3000 steps, particularly for the 3B and 7B models. This indicates that LoRA efficiently adapts the decoder to generate relevant explanations early in the training process.
    \item \textbf{Steady Accuracy Improvement and Plateau:} Predictive accuracy generally improves more steadily over training steps, with most performance gains achieved by the 6000-9000 step mark, after which metrics tend to plateau. This suggests sufficient training duration is needed for the model to learn the failure prediction task effectively.
    \item \textbf{Stable Inference Speed Despite Performance Gains:} Critically, while task performance metrics like Accuracy and ROUGE scores significantly improve during fine-tuning, the inference speed for each model remains largely constant throughout the process (after initial adaptation). This empirically confirms that LoRA achieves performance improvements without introducing additional computational latency at inference time, maintaining the base model's processing efficiency.
\end{enumerate}

These observations from the training process validate key aspects of our approach. The rapid initial gains, especially in language metrics, demonstrate LoRA's efficiency in aligning the VLM to our specific task domain. The stable inference speed confirms the viability of using LoRA for developing real-time capable systems like LKAlert with performance improvement compared to the base model's inference speed.

\subsection{Ablation study: comparison of guided vs.\ unguided}

Table~\ref{tab:lkalert_results} contrasts six model variants, comparing performance with ('Guided') and without ('Unguided') the inclusion of segmentation mask inputs across the three different VLM backbones.

Across all model scales, adding the two segmentation masks (\emph{Guided}) consistently improves primary performance metrics. This enhancement comes with an expected impact on SPS, as processing the additional mask inputs requires more computation compared to the unguided versions. The effectiveness of incorporating this guidance is reflected in the improved average performance ranks achieved by the guided models, particularly for the larger scales (e.g., average rank improving from 3.67 to 2.50 for the 3B model, and from 3.00 to \textbf{1.83} for the 7B model). This highlights the benefit of mask-conditioned attention in eliciting lane-aware reasoning within the VLM.



\section{Conclusion}
\label{sec:conclusion}

We present \textbf{LKAlert}, a real-time, VLM-based early warning system that anticipates LKA failures with interpretable outputs. Through expert model's attention huiding and LoRA fine-tuning, the system enhances safety predictions. Our \textbf{OpenLKA-Alert} dataset bridges the gap between raw perception data and safety-critical decision making, enabling robust binary alert and natural language explanation generation.

Experimental validation confirmed the effectiveness of this framework. LKAlert achieves reliable failure prediction (69.8\% accuracy, 58.6\% F1-score) and high-quality explanation generation (71.7 ROUGE-L), operating efficiently at approximately 2 Hz. This demonstrates its suitability for real-time, resource-constrained deployment to enhance the safety and usability of current ADAS.

Current validation is offline. Future work necessitates real-world deployment and user studies to rigorously assess LKAlert's impact on driver awareness, trust, and response effectiveness in live driving scenarios. This research provides a practical system and a promising methodological framework for advancing human-centered supervision of complex automated driving technologies.



\bibliographystyle{IEEEtran}
\bibliography{references}  

\end{document}